\documentclass{article}
\usepackage{spconf,amsmath,graphicx}
\usepackage{adjustbox}

\usepackage{times}
\usepackage{epsfig}
\usepackage{graphicx}
\usepackage{amsmath}
\usepackage{amssymb}
\usepackage{comment}
\usepackage[ruled,vlined,linesnumbered]{algorithm2e}

\usepackage{times}
\usepackage{epsfig}
\usepackage{graphicx}
\usepackage{amsmath}
\usepackage{amssymb}
\usepackage{adjustbox}
\usepackage{times}
\usepackage{epsfig}
\usepackage{graphicx}
\usepackage{amsmath}
\usepackage{amssymb}

\usepackage[utf8]{inputenc} 
\usepackage[T1]{fontenc}    
\usepackage{url}            
\usepackage{booktabs}       
\usepackage{amsfonts}       
\usepackage{nicefrac}       
\usepackage{bbm}            
\usepackage{multirow}
\usepackage{enumitem}
\usepackage[accsupp]{axessibility}  
\usepackage{float}
\usepackage{microtype}
\usepackage[ruled,vlined,linesnumbered]{algorithm2e}
\usepackage{hyperref}
\usepackage[dvipsnames]{xcolor}

\newcommand{\head}[1]{{\medskip\noindent\textbf{#1}\hspace{10pt}}}

\newcommand{\eq}[1]{(\ref{eq:#1})}

\newcommand{\Th}[1]{\textsc{#1}}
\newcommand{\mr}[2]{\multirow{#1}{*}{#2}}
\newcommand{\mc}[2]{\multicolumn{#1}{c}{#2}}
\newcommand{\tb}[1]{\textbf{#1}}

\newcommand{\red}[1]{{\color{red}{#1}}}

\newcommand{\citeme}[1]{\red{[XX]}}
\newcommand{\refme}[1]{\red{(XX)}}

\newcommand{\tran}{^\top}

\newcommand{\real}{\mathbb{R}}

\newcommand{\mif}{\textrm{if}\ }
\newcommand{\other}{\textrm{otherwise}}

\newcommand{\defn}{\mathrel{:=}}

\newcommand{\cW}{\mathcal{W}}
\newcommand{\cX}{\mathcal{X}}

\newcommand{\vv}{\mathbf{v}}

\newcommand{\vone}{\mathbf{1}}

\makeatletter
\newcommand*\bdot{\mathpalette\bdot@{.7}}
\newcommand*\bdot@[2]{\mathbin{\vcenter{\hbox{\scalebox{#2}{$\m@th#1\bullet$}}}}}
\makeatother

\makeatletter
\DeclareRobustCommand\onedot{\futurelet\@let@token\@onedot}
\def\@onedot{\ifx\@let@token.\else.\null\fi\xspace}

\makeatother

\newcommand{\base}{\mathrm{base}}
\newcommand{\novel}{\mathrm{novel}}

\newcommand{\NN}{\mathrm{NN}}

\newcommand{\aalp}{\mathrm{A^2LP}}

\newcommand{\ci}[1]{{\tiny $\pm$#1}}
\newcommand{\cip}{\phantom{\ci{0.00}}}

\title{Adaptive anchor label propagation for transductive few-shot learning}
\name{Michalis Lazarou$^{1}$ \qquad Yannis Avrithis$^{2}$ \qquad Guangyu Ren$^{1}$ \qquad Tania Stathaki$^{1}$}
  
\address{$^{1}$Department of Electronic and Electronic Engineering, Imperial College London\\
$^{2}$ Institute of Advanced Research on Artificial Intelligence (IARAI)} 
\begin{document}
%
\maketitle
\begin{abstract}
Few-shot learning addresses the issue of classifying images using limited labeled data. Exploiting unlabeled data through the use of transductive inference methods such as label propagation has been shown to improve the performance of few-shot learning significantly. Label propagation infers pseudo-labels for unlabeled data by utilizing a constructed graph that exploits the underlying manifold structure of the data. However, a limitation of the existing label propagation approaches is that the positions of all data points are fixed and might be sub-optimal so that the algorithm is not as effective as possible. In this work, we propose a novel algorithm that adapts the feature embeddings of the labeled data by minimizing a differentiable loss function optimizing their positions in the manifold in the process. Our novel algorithm, \emph{Adaptive Anchor Label Propagation}, outperforms the standard label propagation algorithm by as much as $7\%$ and $2\%$ in the 1-shot and 5-shot settings respectively. We provide experimental results highlighting the merits of our algorithm on four widely used few-shot benchmark datasets, namely \emph{mini}ImageNet, \emph{tiered}ImageNet, CUB and CIFAR-FS and two commonly used backbones, ResNet12 and WideResNet-28-10. The source code can be found at \url{https://github.com/MichalisLazarou/A2LP}.
\end{abstract}
\begin{keywords}
few-shot learning, label propagation, transductive inference
\end{keywords}

\section{Introduction}
\label{sec:intro}


Enabling deep learning models to learn from limited labeled data has attracted significant interest in the research community, evident from the research output in the field of few-shot learning \cite{prototypical, MAML, matchingNets}. Few-shot learning investigates how machine learning models can learn from limited data, for example given only one or a few training examples per class. Multiple approaches have been proposed to address the few-shot learning problem, some of them include meta-learning methods \cite{prototypical, MAML}, representation learning methods \cite{manifoldmixup, rfs} and synthetic data generation methods \cite{TFH, AFHN, VIFSL}.


While most of the aforementioned methods focus on inductive few-shot learning, transductive inference approaches have also been proposed as a way to leverage both labeled training examples and unlabeled test examples, referred to as queries \cite{TPN, iLPC, lrici}. Label propagation \cite{semilpavrithis, lp_ghahramani} is a well known transductive algorithm that has been used as a component in many few-shot transductive methods such as \cite{TPN, iLPC, embeddingpropagation}. Label propagation exploits the data manifold by propagating labels from labeled to unlabeled data along the manifold. However, a limitation of the existing label propagation methods is that the labeled data are fixed and might not be in the optimal position to make the most accurate predictions.

\begin{figure}
\centering
\includegraphics[width=1\linewidth]{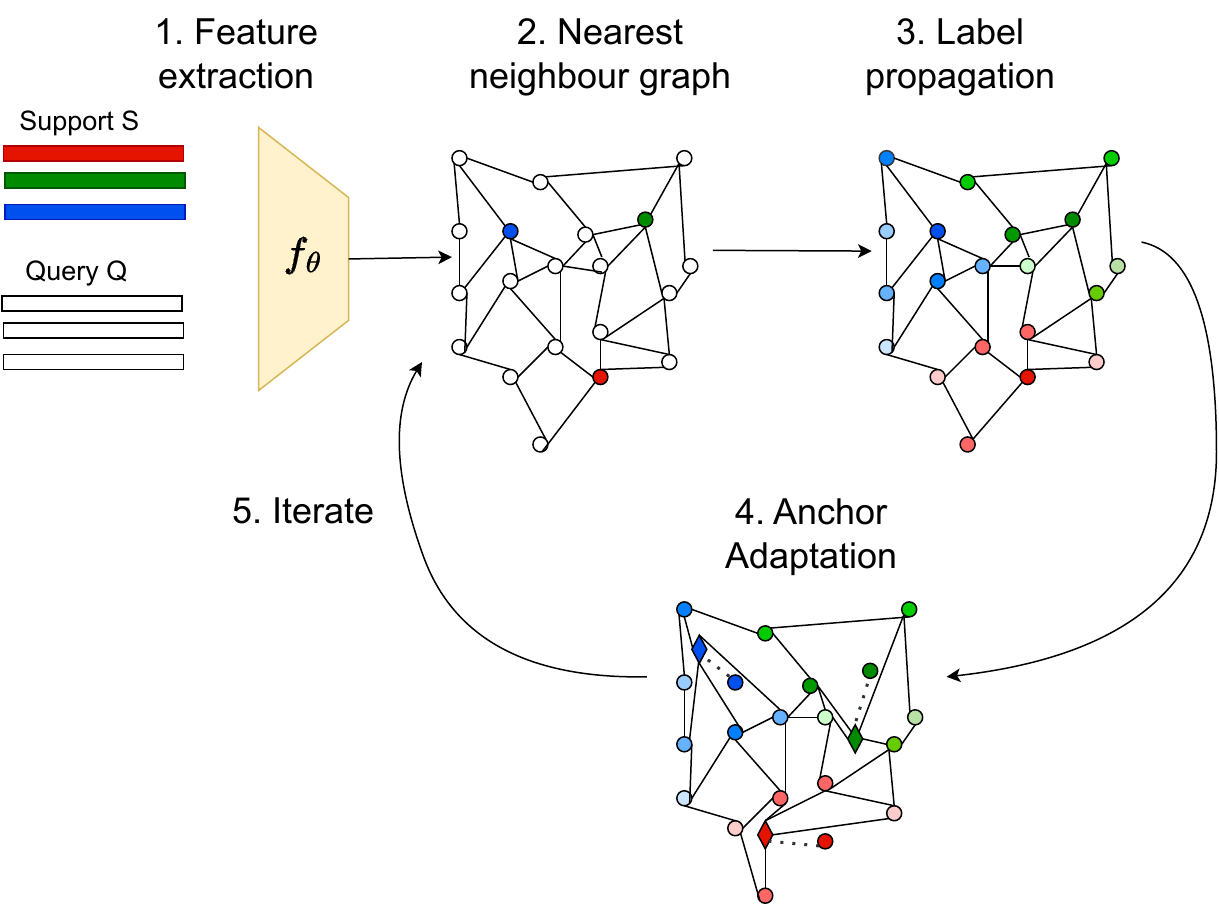}

\caption{Overview of our method. 1) Feature extraction of the support and query sets. 2) $k$-nearest neighbour graph construction. 3) Label propagation from labeled to unlabeled data. 4) Labeled anchor adaptation by minimizing a loss function. The diamonds represent the new position of the anchors and are connected with a dotted line to show the change from their original position. 5) We iterate this procedure over $t$-$\mathrm{steps}$.}
\label{fig:idea}
\end{figure}
To this end we investigate whether optimizing the position of the labeled data, referred as labeled anchors, can improve the performance of label propagation. More specifically we propose a novel algorithm named \emph{Adaptive Anchor Label Propagation} ($\aalp$) in which we adapt the feature embeddings of the labeled anchors by minimizing a multi-class cross entropy loss, placing them in a better position so that the performance of label propagation is improved. 
An overview of our algorithm can be seen in Figure \ref{fig:idea}.

Our contributions are summarized below:
\begin{itemize}
    \item We propose a novel variant of label propagation algorithm named Adaptive Anchor Label Propagation ($\aalp$).
   \item We show the successful application of $\aalp$ in the transductive few-shot learning setting.
    \item $\aalp$ outperforms significantly the standard label propagation as well as multiple state of the art methods on four benchmark datasets using two different backbones.
\end{itemize}

\section{Related work}

\subsection{Few-shot learning}
\label{sec:fsl}
The most commonly researched paradigm addressing few-shot learning is meta-learning.  There are three distinct lines of research in the field of meta-learning: optimization-based meta-learning \cite{MAML, reptile}, metric-based meta-learning \cite{matchingNets, prototypical} and model-based meta-learning \cite{santoro, metaNet}.  Another research direction addressing few-shot learning investigates how to obtain better feature representations utilizing various training techniques to train neural networks such as mixup \cite{manifoldmixup}, self-distillation \cite{rfs} and pretext tasks such as predicting image rotations \cite{rotations}. Synthetic data generation methods have also been prominent in the few-shot learning literature as a way to address the data deficiency of few-shot learning by using synthetic data. These methods exploit generative models such as GANs \cite{AFHN}, VAEs \cite{VIFSL} or other proposed generative models such as \cite{TFH}.

\subsection{Transductive few-shot learning}
\label{sec:transductive_fsl}
Transductive few-shot learning methods utilize both labeled and unlabeled examples during inference to make predictions, achieving remarkable performance improvements over inductive methods. Some ideas that have been proposed in literature include iterative query selection \cite{iLPC, lrici} where rather than making predictions for all the queries at the same time, the most trustworthy queries are selected to augment the labeled dataset. This procedure is iterated until all queries are classified. Other methods utilize the manifold structure of the data through the use of label propagation to train an embedding network \cite{TPN} or to assign pseudo-labels to the queries \cite{iLPC}. Embedding propagation \cite{embeddingpropagation} also exploits the data manifold in order to smoothen the decision boundaries and reduce the noise of the class representations.
Other ideas build on the direction of \cite{prototypical} by exploiting the unlabeled queries to improve the class prototypes, for example by using soft k-means \cite{pt_map} or by selecting the most confident queries per class \cite{crossattention}.

Our work investigates the well-known algorithm of label propagation \cite{semilpavrithis}.  As stated in \ref{sec:intro}, a limitation of the existing label propagation approaches is that the labeled data have fixed positions that might be sub-optimal. To this end, we propose the \emph{Adaptive Anchor Label Propagation} ($\aalp$) algorithm that iteratively adapts the feature embeddings of the labeled anchors, optimizing their positions such that label propagation can provide more accurate predictions.
\section{Method}
\label{sec:method}

\subsection{Problem formulation}
\label{sec:problem_formulation}
At the embedding network pre-training stage, we assume access to a labeled dataset, $D_{\base}$, where every image has a label corresponding to one of the classes from $C_{\base}$ and an embedding network $f_{\theta}$. $D_{\base}$ is used to pre-train the embedding network, $f_{\theta}$, that maps every image from image space $\cX$ to a vector in the embedding space of dimensionality $d$, $f_{\theta}: \cX \to \real^{d}$.

At inference stage, we use the pre-trained embedding network, $f_{\theta}$, to solve novel tasks sampled from a dataset $D_{\novel}$ where every image belongs to a class from $C_{\novel}$. It should be noted that $C_{\base} \cap C_{\novel} = \emptyset$. Following the common few-shot classification setting \cite{matchingNets, prototypical}, we calculate the accuracy performance over numerous novel tasks. An $N$-way $K$-shot novel task consists of a support set, $S$, of $K$ labeled examples per class from $N$ classes sampled from $D_{\novel}$. Therefore, the total number of labeled examples in $S$ is $N_S \defn NK$. In the transductive few-shot learning setting we assume access to the whole query set, $Q$, which consists of all the unlabeled query examples. The query set, $Q$, consists of examples from the same classes as the support set, $S$, with $M$ examples per class. Therefore, the total number of examples in $Q$ is $N_Q \defn NM$. Given $f_{\theta}$, $S$ and $Q$ the goal is to classify all the queries from $Q$ to one of the $N$ classes from $S$.

\subsection{Feature embeddings}
\label{sec:features}
We embed all examples from $S$ and $Q$ using $f_{\theta}$ into the vector set $V \defn \{\vv_i\}^T_{i=1}$ where $\vv_i \defn f_{\theta}(x_i)$ and $T \defn N_S + N_Q$.
We $\ell_2$-normalize the vector set $V$.

\subsection{Nearest neighbour graph construction}
\label{sec:nn_graph}
Following \cite{iLPC}, we construct the $k$-nearest neighbour graph of the vector set $V$ by calculating the sparse affinity matrix, $A$ defined as
\begin{equation}
	A_{ij} \defn
	\begin{cases}
		[\mathrm{cos}(\vv_i,\vv_j)]_{+}^\gamma, & \mif i \ne j \wedge \mathrm{v}_i \in \NN_k(\vv_j) \\
		0,                           & \other
	\end{cases}
\label{eq:affinity}
\end{equation}
where $[\mathrm{cos}(\vv_i,\vv_j)]_{+}^\gamma$ is the nonnegative cosine similarity of vectors $\vv_i$ and $\vv_j$ raised to the power of $\gamma$, where $\gamma$ is a hyperparameter and $\mathrm{NN_k}(\vv_j)$ represents the $k$-nearest neighbours of $\vv_j$ in $\mathrm{V}$. We obtain the adjacency matrix $W \defn \frac{1}{2} (A + A\tran)$ before symmetrically normalizing it as
\begin{equation}
    \cW \defn D^{-1/2} W D^{-1/2},
\label{eq:adj}
\end{equation}
where $D$ is the $T \times T$ degree matrix of $\cW$ defined as
\begin{equation}
    D \defn \mathrm{diag}(\cW \vone_T)
\label{eq:diagonal}
\end{equation}
where $\vone_T$ is the all-ones vector of dimensionality $T$.
\subsection{Label propagation}
\label{sec:lp}
Following \cite{ZBL+03}, we construct the $T \times N$ label matrix $Y$ as
\begin{equation}
 Y_{ij} \defn
	\begin{cases}
		1, & \mif i \in N_S \wedge y_i=j\\
		0, & \other
	 \end{cases}
\label{eq:label}
\end{equation}
In other words, the first $N_S$ rows of $Y$ represent the one-hot labels of the support set, $S$, while the rest of the rows represent the query set $Q$ and are filled with zeros. Label propagation is defined as
\begin{equation}
    Z \defn (I-\alpha \cW)^{-1} Y,
\label{eq:lp}
\end{equation}
where $I$ is the identity matrix and $\alpha \in [0, 1)$ is a hyperparameter. By solving equation \eq{lp} we obtain the manifold similarity matrix $Z$, where the row vector $z_i$ at row $i$ expresses how similar example $\vv_i$ is to each of the $N$ support classes. 

\subsection{Adapting labeled anchors}
\label{sec:adapting_anchors}
We leverage the manifold similarity matrix $Z$ in order to adapt the feature embeddings of the support vectors which serve as the labeled anchors, $V_S \defn \{\vv_i\}^{N_S}_{i=1}$, and optimize their position in the k-nearest neighbour graph.  
We obtain the labeled anchor probability matrix $P^{a}$ by calculating the probability distribution for every labeled anchor $\vv_i \in V_S$ using the softmax function. Therefore the probability distribution $p^a_i$ for the labeled anchor $\vv_i$ is defined as: 

\begin{equation}
    p^a_{i} \defn \frac{\exp(\tau z_{i})}{\sum_{j=1}^N \exp(\tau z_{ij})},
\label{eq:softmax}
\end{equation}
where $\tau$ is a scalar and $N$ is the total number of support classes. 
Using \eq{softmax} we calculate the cross-entropy loss with respect to set $V_S$
\begin{equation}
   L_{ce}(V_S) \defn -\displaystyle\sum_{i \in N_S} \sum_{j \in N}y_{ij}\log(p^a_{ij}),
\label{eq:ce}
\end{equation}
where $y_{ij}$ denotes the $j$-th component of the one-hot label $y_i$ associated with $\vv_i$ and $p^a_{ij}$ denotes the $j$-th component of $p^a_i$. We update the feature embeddings of the labeled anchors $V_S$ by minimizing \eq{ce} using gradient descent or any other gradient-based optimization algorithm with respect to $V_S$. 


\subsection{Iteration and Inference}
\label{sec:iteration}
We iterate the process from graph construction to adapting the labeled anchors for $t$-$\mathrm{steps}$. 
Upon completion of the iteration process, we use the final $Z$ to classify every query in $Q$ to the class with the highest manifold class similarity \cite{ZBL+03}. The predicted labeled set of $Q$ is denoted as $\hat{Y_Q}$.

Algorithm \ref{alg:adaptive_lp} summarizes our method for a single task.


\begin{algorithm}
\footnotesize
\DontPrintSemicolon
\SetFuncSty{textsc}
\SetDataSty{emph}
\newcommand{\commentsty}[1]{{\color{Blue}#1}}
\SetCommentSty{commentsty}
\SetKwComment{Comment}{$\triangleright$ }{}
\SetKwInOut{Input}{input}
\SetKwInOut{Output}{output}
\SetKwFunction{Graph}{graph}
\SetKwFunction{Label}{label}
\SetKwFunction{Power}{power}
\SetKwFunction{Balance}{balance}
\SetKwFunction{Sinkhorn}{Sinkhorn}
\SetKwFunction{Predict}{predict}
\SetKwFunction{Clean}{clean}
\SetKwFunction{Augment}{augment}
\SetKwComment{Comment}{$\triangleright$ }{}

\SetKwInOut{Input}{input}
\SetKwInOut{Output}{output}
\SetKwFunction{Graph}{graph}
\SetKwFunction{Label}{label}
\SetKwFunction{Softmax}{softmax}
\SetKwFunction{Ce}{ce}

\SetKwFunction{LP}{lp}
\SetKwFunction{Update}{update}
\SetKwFunction{Power}{power}
\SetKwFunction{Balance}{balance}
\SetKwFunction{Sinkhorn}{Sinkhorn}
\SetKwFunction{Predict}{predict}
\SetKwFunction{Clean}{clean}
\SetKwFunction{Augment}{augment}

\Input{ Pre-trained $f_{\theta}$}
\Input{ Support set, $S$}
\Input{ Query set, $Q$}
\Output{ Predicted labels for $Q$, $\hat{Y}_Q$}
\BlankLine

Calculate $V$ by embedding $S$ and $Q$ using $f_{\theta}$\\
$\ell_2$-normalize every vector $\vv$ in $V$\\
\For{$t$-$\mathrm{steps}$}
{
    $\cW \gets \Graph(f_{\theta},S,Q;\gamma,k)$ \Comment*{Adjacency matrix~\eq{affinity},\eq{adj}} 
	$Y \gets \Label(S)$ \Comment*{Label matrix~\eq{label}}
	$Z \gets \LP(\cW,Y;\alpha)$ \Comment*{Label propagation~\eq{lp}}
	$P^a \gets \Softmax(Z[0:N_S])$ \Comment*{Probability matrix~\eq{softmax}}
	$L_{ce}(V_S) \gets \Ce(P^a, Y[0:N_S])$ \Comment*{Cross entropy loss~\eq{ce}}
    $V_S \gets \Update(V_S, L_{ce}(V_S))$ \Comment*{Update labeled anchors}
}
Predict $\hat{Y_Q}$
\caption{Adaptive Anchor Label Propagation}
\label{alg:adaptive_lp}
\end{algorithm}

\section{Experiments}

\begin{table*}
\small
\centering
\setlength\tabcolsep{4pt}
\begin{tabular}{lcccccccccc} \toprule
\mr{1}{\Th{Algorithm}}                            & \mc{2}{\Th{\emph{mini}ImageNet}} & \mc{2}{\Th{\emph{tiered}ImageNet}}          & \mc{2}{\Th{CIFAR-FS}}                          & \mc{2}{\Th{CUB}}                    \\ \cmidrule{2-9}
                                              &     1-shot           & 5-shot       & 1-shot              & 5-shot              & 1-shot              & 5-shot              & 1-shot              & 5-shot        \\ \midrule

\mc{9}{\Th{ResNet-12}}   \\ \midrule
Prototypical classifier          & 54.23\ci{0.61}   & 74.98\ci{0.48}    & 67.21\ci{0.72} & 85.19\ci{0.48} & 60.65\ci{0.71}  & 80.45\ci{0.52}  &  75.13\ci{0.62} & 89.78\ci{0.38}    \\
Imprinting+$L_{ce}$           & 54.53\ci{0.61}   & 74.31\ci{0.49}    & 67.53\ci{0.72}  & 84.86\ci{0.48} & 61.12\ci{0.69}  & 80.32\ci{0.53}  &  75.03\ci{0.62} & 89.49\ci{0.37}    \\
LP            & 59.15\ci{0.66}   & 73.50\ci{0.52}   & 72.67\ci{0.74}   & 84.61\ci{0.53}   & 66.60\ci{0.74}   & 80.47\ci{0.53}  & 79.73\ci{0.61}   & 89.51\ci{0.39}   \\
$\aalp$         & \tb{64.35}\ci{0.77}   & \tb{75.00}\ci{0.53}   & \tb{80.49}\ci{0.77}   & \tb{85.80}\ci{0.51}   & \tb{73.27}\ci{0.81}   & \tb{81.32}\ci{0.55}   & \tb{87.15}\ci{0.58}   & \tb{90.35}\ci{0.37}     \\
\midrule
\mc{9}{\Th{WideResNet-28-10}}   \\ \midrule  

Prototypical classifier          & 65.35\ci{0.63}   & 83.37\ci{0.43}    & 73.47\ci{0.70}  & 88.22\ci{0.45} & 74.14\ci{0.67}  & 87.05\ci{0.47}  &  81.06\ci{0.64} & 90.82\ci{0.33}    \\
Imprinting+$L_{ce}$           & 66.07\ci{0.62}   & 83.34\ci{0.42}    & 74.07\ci{0.69}  & 88.55\ci{0.43} & 74.16\ci{0.66}  & 87.15\ci{0.47}  &  81.10\ci{0.63} & 91.17\ci{0.33}    \\
LP            & 69.50\ci{0.64}   & 81.28\ci{0.47}   & 78.14\ci{0.72}  & 87.63\ci{0.50}  &  78.66\ci{0.67}  & 87.19\ci{0.51}  & 85.11\ci{0.63}  & 91.37\ci{0.36}    \\
$\aalp$        & \tb{76.48}\ci{0.75}  & \tb{83.57}\ci{0.45}   &  \tb{82.82}\ci{0.73}  &  \tb{88.80}\ci{0.46}  & \tb{82.15}\ci{0.71}  & \tb{87.95}\ci{0.48} &  \tb{86.45}\ci{0.65}  & \tb{91.66}\ci{0.34}     \\ 
\bottomrule

\end{tabular}
\vspace{6pt}
\caption{\emph{Transductive inference}. Comparisons of implemented baselines.}
\label{tab:wrn}
\end{table*}

    
\begin{table}
\small
\centering
\begin{adjustbox}{width=1\columnwidth}
\setlength\tabcolsep{4pt}
\begin{tabular}{lcccccc} \toprule
\mr{1}{\Th{Algorithm}}                            & \mc{2}{\Th{\emph{mini}ImageNet}} & \mc{2}{\Th{\emph{tiered}ImageNet}}                 \\ \cmidrule{2-5}
                                              &     1-shot           & 5-shot       & 1-shot              & 5-shot        \\ \midrule
\mc{5}{\Th{WideResNet-28-10}}   \\ \midrule  
Prototypical classifier        & 69.64\ci{0.60}   & 84.61\ci{0.42}    & 77.26\ci{0.65}  & 89.22\ci{0.42}     \\
Imprint+$L_{ce}$           & 68.77\ci{0.60}   & 84.24\ci{0.42}    & 76.13\ci{0.66}  & 88.95\ci{0.42}    \\
LP              & 74.24\ci{0.66}   & 84.59\ci{0.44}    & 82.48\ci{0.70}  & 90.07\ci{0.45}  \\
$\aalp$        & \tb{75.94}\ci{0.72}  & \tb{85.67}\ci{0.42}   &  \tb{83.68}\ci{0.72}  & \tb{90.53}\ci{0.43}    \\
\bottomrule

\end{tabular}
\end{adjustbox}
\vspace{6pt}
\caption{\emph{Transductive inference using PLC pre-processing}. Comparisons of implemented baselines.}
\label{tab:plc}
\end{table}

\subsection{Setup}

\head{Datasets}
We conduct exeperiments on four widely used few-shot learning benchmark datasets. These are: \emph{mini}Imagenet \cite{matchingNets}, \emph{tiered}Imagenet \cite{manifoldmixup}, CUB \cite{closerlook} and CIFAR-FS \cite{closerlook}.

\head{Tasks} We follow the paradigm of $N$-way, $K$-shot learning, where a task consists of a support set, $S$, of $N=5$ classes and $K\in \{1, 5\}$ labeled examples per class. A task also includes a query set $Q$ with $M=15$ queries per class, making the total number of queries $N_Q = NM = 75$. We report the mean accuracy and the 95$\%$ confidence interval over 1000 sampled tasks from the novel dataset $D_{\novel}$.

\head{Networks}
We experiment with two different network architectures, namely ResNet-12 and WideResNet-28-10. We use the publicly available pre-trained weights of both networks provided from \footnote{\url{https://github.com/MichalisLazarou/iLPC}}. ResNet-12 was trained following \cite{lrici} and WideResNet-28-10 was trained following \cite{manifoldmixup}.

\head{Implementation details and hyperparameters} Our implementation is based on python using the PyTorch framework \cite{pytorch}. We set $\gamma=3$ in \eq{affinity}, $\tau=15$ in \eq{softmax}, $\alpha=0.8$ in \eq{lp} and $k=20$ in \eq{affinity}. We set $t$-$\mathrm{steps}=1000$ in section \ref{sec:iteration} and use Adam optimizer with a learning rate $\eta=0.0001$. We keep our hyperparameters fixed under all different datasets, backbones, 1-shot and 5-shot settings in order to keep our experiments simple and avoid overfitting.

\subsection{Experimental results}
\head{Baselines}
In order to have fair and objective experimental comparisons, we implement the following inference methods to serve as baselines: 1) the widely used prototypical classifier \cite{prototypical}, 2) the weight imprinting method \cite{imprintedweights} followed by fine-tuning using $L_{ce}$ (Imprinting+$L_{ce}$) and 3) the standard label propagation (LP) as it was used in \cite{iLPC, semilpavrithis}. 

\head{Discussion}
It can be seen from Table \ref{tab:wrn} that $\aalp$ outperforms significantly all other baselines in both 1-shot and 5-shot settings under all datasets and backbones. Impressively, $\aalp$ outperforms the standard label propagation (LP) by more than $7\%$ in the 1-shot CUB using ResNet-12 as shown in Table \ref{tab:wrn}, highlighting the importance of adapting the labeled anchors' feature embeddings. 
Interestingly, simply fine-tuning the imprinted weights using $L_{ce}$ (Imprinting+$L_{ce}$) cannot provide the same performance increase as $\aalp$ because $\aalp$ exploits the node connections of the constructed graph to adapt the labeled anchors in contrast to Imprint+$L_{ce}$.

We also investigate the robustness of $\aalp$ using the feature pre-processing technique proposed in \cite{iLPC, pt_map}, referred to as PLC pre-processing. Instead of using only $\ell_2$-normalization, PLC pre-processing consists of element-wise power transform $\vv^{\frac{1}{2}}$ for $\vv \in V$, followed by $\ell_2$-normalization and centering by subtracting the mean of all vectors in $V$. It can be seen from Table \ref{tab:plc} that $\aalp$ still outperforms all other baselines.

Lastly, we compare $\aalp$ with several state of the art transductive few-shot learning methods. 
Since each method uses different training regimes and network architectures, in order to make the comparisons as fair as possible we compare the best version of every method including ours. It can be seen from Table \ref{tab:sota} that $\aalp$ outperforms the other methods in 3 out of 4 settings, outperforming its closest competitor by $\sim2\%$ in the 1-shot \emph{tiered}Imagenet setting. Furthermore, either $\aalp$ or $\aalp$ with PLC pre-processing outperforms all competitors.


\begin{table}
\small
\centering
\begin{adjustbox}{width=1\columnwidth}
\setlength\tabcolsep{4pt}
\begin{tabular}{lcccccc} \toprule
\mr{1}{\Th{Algorithm}}                            & \mc{2}{\Th{\emph{mini}ImageNet}} & \mc{2}{\Th{\emph{tiered}ImageNet}}                 \\ \cmidrule{2-5}
                                               &     1-shot           & 5-shot       & 1-shot              & 5-shot        \\ \midrule
                                            \mc{5}{\Th{ResNet-12}}   \\ \midrule  
   
LR+ICI~\cite{lrici}    & 66.80\cip           & 79.26\cip           & 80.79\cip           & 87.92\cip    \\
CAN+Top-\emph{k}~\cite{crossattention} & 67.19\ci{0.55}      & 80.64\ci{0.35}      & 73.21\ci{0.58}      & 84.93 \ci{0.38} \\
DPGN~\cite{DPGN} & 67.77\ci{0.32}      & 84.60\ci{0.43}      & 72.45\ci{0.51}      & 87.24\ci{0.39}      \\      
\midrule \mc{5}{\Th{WideResNet-28-10}}   \\ \midrule  
EP \cite{embeddingpropagation} & 70.74\ci{0.85} & 84.34\ci{0.53} & 78.50\ci{0.91} & 88.36\ci{0.57}\\
SIB \cite{SIB} & 70.00\ci{0.60} & 79.20\ci{0.40}& 72.90\cip & 82.80\cip \\
SIB+E$^3$BM~\cite{ensemblefsl}         & 71.40\ci{0.50}      & 81.20\ci{0.40}      & 75.60\ci{0.60}       & 84.30\ci{0.40}\\
LaplacianShot \cite{laplacianshot} & 74.86\ci{0.19} & 84.13\ci{0.14} & 80.18\ci{0.21} & 87.56\ci{0.15} \\
$\aalp$       & \tb{76.48}\ci{0.75}  & 83.57\ci{0.45}   &  82.82\ci{0.73}  &  88.80\ci{0.46}     \\ 
$\aalp$+PLC     &  75.94\ci{0.72}  & \tb{85.67}\ci{0.42}   &  \tb{83.68}\ci{0.72}  & \tb{90.53}\ci{0.43}      \\ 
\bottomrule

\end{tabular}
\end{adjustbox}
\vspace{6pt}
\caption{\emph{Transductive inference}. Comparisons of state of the art methods.}
\label{tab:sota}
\end{table}

\section{Conclusions}
\label{sec:conclusion}
In this work we propose the novel algorithm Adaptive Anchor Label Propagation ($\aalp$). Our algorithm significantly outperforms the label propagation as it was used in \cite{iLPC, semilpavrithis} as well as several state of the art methods in the transductive few-shot learning setting. Through our investigation it is evident that optimizing the position of the labeled anchors is essential for improving the performance of label propagation.  In the future, we are interested to investigate different loss functions for adapting the feature embeddings of the labeled anchors. 

\bibliographystyle{IEEEbib}
\bibliography{refs}

\begin{thebibliography}{10}

\bibitem{prototypical}
Jake Snell, Kevin Swersky, and Richard Zemel,
\newblock ``Prototypical networks for few-shot learning,''
\newblock in {\em NeurIPS}, 2017.

\bibitem{MAML}
Chelsea Finn, Pieter Abbeel, and Sergey Levine,
\newblock ``Model-agnostic meta-learning for fast adaptation of deep
  networks,''
\newblock in {\em ICML}, 2017.

\bibitem{matchingNets}
Oriol Vinyals, Charles Blundell, Timothy Lillicrap, Daan Wierstra, et~al.,
\newblock ``Matching networks for one shot learning,''
\newblock in {\em NeurIPS}, 2016.

\bibitem{manifoldmixup}
Puneet Mangla, Nupur Kumari, Abhishek Sinha, Mayank Singh, Balaji
  Krishnamurthy, and Vineeth~N Balasubramanian,
\newblock ``Charting the right manifold: Manifold mixup for few-shot
  learning,''
\newblock in {\em WACV}, 2020.

\bibitem{rfs}
Yonglong Tian, Yue Wang, Dilip Krishnan, Joshua~B Tenenbaum, and Phillip Isola,
\newblock ``Rethinking few-shot image classification: a good embedding is all
  you need?,''
\newblock {\em arXiv preprint arXiv:2003.11539}, 2020.

\bibitem{TFH}
Michalis Lazarou, Tania Stathaki, and Yannis Avrithis,
\newblock ``Tensor feature hallucination for few-shot learning,''
\newblock in {\em WACV}, 2022.

\bibitem{AFHN}
Kai Li, Yulun Zhang, Kunpeng Li, and Yun Fu,
\newblock ``Adversarial feature hallucination networks for few-shot learning,''
\newblock in {\em CVPR}, 2020.

\bibitem{VIFSL}
Qinxuan Luo, Lingfeng Wang, Jingguo Lv, Shiming Xiang, and Chunhong Pan,
\newblock ``Few-shot learning via feature hallucination with variational
  inference,''
\newblock in {\em WACV}, 2021.

\bibitem{TPN}
Y~Liu, J~Lee, M~Park, S~Kim, E~Yang, SJ~Hwang, and Y~Yang,
\newblock ``Learning to propagate labels: Transductive propagation network for
  few-shot learning,''
\newblock in {\em ICLR}, 2019.

\bibitem{iLPC}
Michalis Lazarou, Tania Stathaki, and Yannis Avrithis,
\newblock ``Iterative label cleaning for transductive and semi-supervised
  few-shot learning,''
\newblock in {\em ICCV}, 2021.

\bibitem{lrici}
Yikai Wang, C.~Xu, Chen Liu, Liyong Zhang, and Yanwei Fu,
\newblock ``Instance credibility inference for few-shot learning,''
\newblock {\em CVPR}, 2020.

\bibitem{semilpavrithis}
Ahmet Iscen, Giorgos Tolias, Yannis Avrithis, and Ondrej Chum,
\newblock ``Label propagation for deep semi-supervised learning,''
\newblock in {\em CVPR}, 2019.

\bibitem{lp_ghahramani}
Xiaojin Zhu and Zoubin Ghahramani,
\newblock ``Learning from labeled and unlabeled data with label propagation,''
\newblock Tech. {R}ep., 2002.

\bibitem{embeddingpropagation}
Pau Rodr{\'\i}guez, Issam Laradji, Alexandre Drouin, and Alexandre Lacoste,
\newblock ``Embedding propagation: Smoother manifold for few-shot
  classification,''
\newblock in {\em ECCV}, 2020.

\bibitem{reptile}
Alex Nichol, Joshua Achiam, and John Schulman,
\newblock ``On first-order meta-learning algorithms,''
\newblock {\em arXiv preprint arXiv:1803.02999}, 2018.

\bibitem{santoro}
Adam Santoro, Sergey Bartunov, Matthew Botvinick, Daan Wierstra, and Timothy
  Lillicrap,
\newblock ``Meta-learning with memory-augmented neural networks,''
\newblock in {\em ICML}, 2016.

\bibitem{metaNet}
Tsendsuren Munkhdalai and Hong Yu,
\newblock ``Meta networks,''
\newblock in {\em ICML}, 2017.

\bibitem{rotations}
Spyros Gidaris, Praveer Singh, and Nikos Komodakis,
\newblock ``Unsupervised representation learning by predicting image
  rotations,''
\newblock in {\em ICLR}, 2018.

\bibitem{pt_map}
Yuqing Hu, Vincent Gripon, and St{\'e}phane Pateux,
\newblock ``Leveraging the feature distribution in transfer-based few-shot
  learning,''
\newblock in {\em International Conference on Artificial Neural Networks}.
  Springer, 2021.

\bibitem{crossattention}
Ruibing Hou, Hong Chang, MA~Bingpeng, Shiguang Shan, and Xilin Chen,
\newblock ``Cross attention network for few-shot classification,''
\newblock in {\em NeurIPS}, 2019.

\bibitem{ZBL+03}
Dengyong Zhou, Olivier Bousquet, Thomas~Navin Lal, Jason Weston, and Bernhard
  Sch{\"o}lkopf,
\newblock ``Learning with local and global consistency,''
\newblock in {\em NIPS}, 2003.

\bibitem{closerlook}
Wei-Yu Chen, Yen-Cheng Liu, Zsolt Kira, Yu-Chiang Wang, and Jia-Bin Huang,
\newblock ``A closer look at few-shot classification,''
\newblock in {\em ICLR}, 2019.

\bibitem{pytorch}
Adam Paszke, Sam Gross, Soumith Chintala, Gregory Chanan, Edward Yang, Zachary
  DeVito, Zeming Lin, Alban Desmaison, Luca Antiga, and Adam Lerer,
\newblock ``Automatic differentiation in pytorch,''
\newblock 2017.

\bibitem{imprintedweights}
Hang Qi, Matthew Brown, and David~G Lowe,
\newblock ``Low-shot learning with imprinted weights,''
\newblock in {\em CVPR}, 2018.

\bibitem{DPGN}
Ling Yang, Liangliang Li, Zilun Zhang, Xinyu Zhou, Erjin Zhou, and Yu~Liu,
\newblock ``Dpgn: Distribution propagation graph network for few-shot
  learning,''
\newblock in {\em CVPR}, 2020.

\bibitem{SIB}
Shell~Xu Hu, Pablo~Garcia Moreno, Yang Xiao, Xi~Shen, Guillaume Obozinski, Neil
  Lawrence, and Andreas Damianou,
\newblock ``Empirical bayes transductive meta-learning with synthetic
  gradients,''
\newblock in {\em ICLR}, 2019.

\bibitem{ensemblefsl}
Yaoyao Liu, Bernt Schiele, and Qianru Sun,
\newblock ``An ensemble of epoch-wise empirical bayes for few-shot learning,''
\newblock in {\em ECCV}, 2020.

\bibitem{laplacianshot}
Imtiaz Ziko, Jose Dolz, Eric Granger, and Ismail~Ben Ayed,
\newblock ``Laplacian regularized few-shot learning,''
\newblock in {\em ICML}, 2020.

\end{thebibliography}

\end{document}